\documentclass[a4paper,times,conference]{IEEEtran}
\IEEEoverridecommandlockouts
\usepackage{cite}
\usepackage{amsmath,amssymb,amsfonts}
\usepackage{graphicx}
\usepackage{textcomp}
\usepackage{xcolor}
\def\BibTeX{{\rm B\kern-.05em{\sc i\kern-.025em b}\kern-.08em
    T\kern-.1667em\lower.7ex\hbox{E}\kern-.125emX}}

\usepackage{booktabs}
\usepackage[disable]{todonotes}
\usepackage[normalem]{ulem}
\usepackage{amssymb,amsfonts}
\usepackage[inline]{enumitem}
\usepackage{multirow}
\usepackage{subcaption}
\usepackage{stfloats}
\usepackage{algorithm}
\usepackage{algpseudocodex}
\usepackage{orcidlink}

\newcommand{\method}{Self-supervised Adversarial Imitation Learning}
\newcommand{\abbrev}{SAIL}

\newcommand{\ie}{{\it i.e.}}
\newcommand{\eg}{{\it e.g.}}

\newcommand{\etal}{{\it et al.}}

\makeatletter
\newcommand{\linebreakand}{%
  \end{@IEEEauthorhalign}
  \hfill\mbox{}\par
  \mbox{}\hfill\begin{@IEEEauthorhalign}
}
\makeatother

\begin{document}

\title{Self-Supervised Adversarial Imitation Learning}

\author{
    \IEEEauthorblockN{
        1\textsuperscript{st} Juarez Monteiro$^1$~\orcidlink{0000-0002-8831-5343}
    }
    \IEEEauthorblockA{
    \textit{Pontif{\'i}cia Universidade Cat{\'o}lica do RS} \\
    Porto Alegre, Brazil \\
    juarez.santos@acad.pucrs.br}
    \and
    \IEEEauthorblockN{
        2\textsuperscript{nd} Nathan Gavenski$^1$~\orcidlink{0000-0003-0578-3086}
    }
    \IEEEauthorblockA{
    \textit{King's College London}\\
    London, England \\
    nathan.schneider\_gavenski@kcl.ac.uk}
    \linebreakand
    \IEEEauthorblockN{
        3\textsuperscript{rd} Felipe Meneguzzi~\orcidlink{0000-0003-3549-6168}
    }
    \IEEEauthorblockA{
    \textit{University of Aberdeen} \\
    \textit{Pontif{\'i}cia Universidade Cat{\'o}lica do RS} \\
    Aberdeen, Scotland \\
    felipe.meneguzzi@abdn.ac.uk}
    \and
    \IEEEauthorblockN{
        4\textsuperscript{th} Rodrigo C. Barros~\orcidlink{0000-0002-0782-9482}
    }
    \IEEEauthorblockA{
    \textit{Pontif{\'i}cia Universidade Cat{\'o}lica do RS}\\
    Porto Alegre, Brazil \\
    rodrigo.barros@pucrs.br}
}

\maketitle

\footnotetext[1]{These authors contributed equally to
the work.}

\begin{abstract}
Behavioural cloning is an imitation learning technique that teaches an agent how to behave via expert demonstrations.
Recent approaches use self-supervision of fully-observable unlabelled snapshots of the states to decode state pairs into actions.
However, the iterative learning scheme employed by these techniques is prone to get trapped into bad local minima.
Previous work uses goal-aware strategies to solve this issue. However, this requires manual intervention to verify whether an agent has reached its goal.
We address this limitation by incorporating a discriminator into the original framework, offering two key advantages and directly solving a learning problem previous work had.
First, it disposes of the manual intervention requirement. 
Second, it helps in learning by guiding function approximation based on the state transition of the expert's trajectories.
Third, the discriminator solves a learning issue commonly present in the policy model, which is to sometimes perform a `no action' within the environment until the agent finally halts.
\end{abstract}


\section{Introduction} \label{sec:introduction}
Learning by observing is an intrinsic human ability that we have been able to rely on since childhood.
We can learn tasks by watching a video teaching us how to cook or how to play a specific video game.
Learning from demonstrations allows humans to learn tasks from proficient sources and apply the newly-acquired knowledge in different domains, similar tasks, or after adapting it to their own reality, \ie, different body sizes and proportions.
Occasionally, learning a task by observing a specialist can be difficult.
We can watch tennis players performing their best moves, but it is not a simple task for us to break down their actions into straightforward instructions to learn them properly.

In Machine Learning~(ML), we refer to the technique of learning from a teacher as Imitation Learning (IL).
It consists of an agent learning from the actions of a known teacher in order to solve a given task~\cite{HusseinEtAl2017}.
The learning agent must be able to achieve the goal or conclude the task which it was trained for.
Recent approaches try to approximate the human learning experience by exploring a strategy where the agent needs no explicit label of the actions performed by the teacher to imitate them.
This strategy of learning without explicitly receiving the teacher's actions (labels) or learning by observing is called in the literature \textit{Learning from Observation} (LfO)~\cite{LiuEtAl2018,TorabiEtAl2018,MonteiroEtAl2020abco,gavenski2020imitating}.

LfO emerges with improvements in efficiency~\cite{zhu2020off} and generalisation~\cite{MonteiroEtAl2020abco}, overcoming the need for fine-grained information found in annotated trajectories or in complex reward functions.
Since they are more effective, LfO methods require fewer teacher snapshots, which helps mitigate the lack-of-labels problem in the available training data.
The fact that we can use LfO with smaller amounts of (unsupervised) data gives us the opportunity to explore problems where data is scarce or costly to collect, \eg, autonomous vehicles~\cite{le2022survey}.
The existing methods for LfO often rely on learning how to map the state-action transition in a self-supervised manner.
This, in turn, requires access to the test environment and multiple deliberations from the agent, so we properly map the state-action transitions.

LfO strategies are frequently benchmarked by measuring \textit{performance} and \textit{efficiency} following a specific formalisation~\cite{TorabiEtAl2018,MonteiroEtAl2020abco,gavenski2020imitating}. 
One can also evaluate imitation by comparing the trajectories that both the agent and the teacher have taken in order to solve a given task.
Both perspectives have their issues. 
With the traditional measures, an agent can diverge from the teacher and still achieve the same reward, even though it is performing in a completely different way from what was expected.
Indeed, sparse rewards in an environment make identifying proficient behaviour non-trivial. 
By only comparing rewards and not trajectories or intent, an agent might reach the same reward as its proficient counterpart, though perhaps missing part of the desired behaviour that it should account for.
We can derive a similar example for the second perspective.
Suppose the agent is acting within a maze environment, where the agent and the teacher follow practically the same trajectory, but in the end, the agent does not achieve a final state, resulting in a totally different reward.

\section{Related Work} \label{sec:related-work}
% -----------------------------------------------
% \subsection{Behavior Cloning (BC)}
% -----------------------------------------------
Behavioural Cloning (BC)~\cite{Pomerleau1988} is one of the most straightforward techniques for Imitation Learning~\cite{zheng2022imitation}. 
BC uses teacher trajectories containing the state $s_t$ and the action $a_t$ of a given task at time $t$ to create a policy $\pi = P(a \mid s_t)$.
BC presents consistent results in the aspect of episodic rewards by training a policy in a supervised manner.
However, it comes with a high cost for wide state-space scenarios, requiring sufficient trajectories to make the policy generalise for unknown states.

% -----------------------------------------------
% \subsection{Behavior Cloning from Observation (BCO)}
% -----------------------------------------------
To solve this issue, recent approaches in IL~\cite{TorabiEtAl2018,TorabiEtAl2019generative,MonteiroEtAl2020abco,gavenski2020imitating} employ strategies that do not require labelled data provided by a teacher.
Torabi~\etal~\cite{TorabiEtAl2018} designed a LfO model-based strategy called Behavioural Cloning from Observation (BCO), which learns to imitate using a self-supervised strategy that does not require a teacher to provide annotated data.
BCO starts by learning the state-action transition to build a predictor capable of guessing what action occurred in a given pair of states $S_t$ and $S_{t+1}$. 
It then uses such a model to label the teacher trajectories.
Next, all automatically-labelled data is used to train a policy in a supervised fashion.
Even though the authors present better results than supervised methods that use the teacher's original labels, the approach lacks a proper exploration technique, leading to a situation in which it repeatedly finds itself stuck in endless states.

Imitating Unknown Policies via Exploration (IUPE)~\cite{gavenski2020imitating} is an approach that uses sampling and exploration mechanisms to solve the efficiency issue.
After each interaction, the inverse dynamics model $\mathcal{M}$ learns the most likely action given a state-transition in a supervised fashion, and then it trains a policy with the pseudo-labelled ($\mathcal{M}$'s outputs) as teacher's actions.
IUPE weighs the random and policy samples to create a dataset similar to the teacher's transitions.
Finally, its exploration mechanism uses the softmax distribution over its outputs to perform weighed sampling over all actions from both models.
By avoiding the usage of its \textit{maximum a posteriori} (MAP) estimation, it creates a stochastic policy that dynamically changes its exploration ratio according to each model's output. 
The downside of using IUPE is its need for hand-crafted goal-aware functions, which require prior domain knowledge, to retrieve intermediate samples capable of approximating the initial random state-action pairs to the proficient ones.

% -----------------------------------------------
% \subsection{Generative Adversarial Imitation Learning (GAIL)}
% -----------------------------------------------
Ho and Ermon~\cite{HoErmon2016} propose Generative Adversarial Imitation Learning (GAIL), an approach that uses adversarial training to solve that same problem. 
GAIL requires a smaller teacher dataset compared to other approaches. Nevertheless, the method needs extensive interactions with the environment.
% -----------------------------------------------
% \subsection{Generative Adversarial Imitation from Observation (GAIfO)}
% -----------------------------------------------
Torabi~\etal~\cite{TorabiEtAl2019generative} find inspiration on GAIL to design a Generative Adversarial Network (GAN)~\cite{GoodfellowEtAl2014} named Generative Adversarial Imitation from Observation (GAIfO).
GAIfO tries to learn a policy by creating a mechanism to distinguish if the source of the data is from a teacher or provided by the model.
By training the model to understand what is the next state $S_{t+1}$ from a given state-action pair $S_t$ and $a$, GAIfO produces a policy that has similar behaviour to a teacher.
Although GAIfO has yielded significantly better results than GAIL, by reducing the number of samples necessary to train a policy, it falls in the same issue as GAIL, where the number of interactions with the environment is a bottleneck.

\section{Problem Formulation} \label{sec:problem}

We formalise Imitation Learning assuming an environment defined by a Markov Decision Process (MDP), which is represented by a five-tuple $M = \{S, A, T, r, \gamma\}$~\cite[Ch 3.]{SuttonBarto2018}, where  $S$ is the state-space, $A$ is the action space, $T$ is the transition model, $r$ is the immediate reward function, and $\gamma$ is the discount factor.
Solving an MDP yields a stochastic policy $\pi(a|s)$ with a probability distribution over actions for an agent in state $s$ that needs to take a given action $a$.
\emph{Imitation from observation} (IfO)~\cite{TorabiEtAl2018} aims to learn the inverse dynamics of the agent, $\mathcal{M}_{a}^{s_t,s_{t+1}} = P(a|s_t,s_{t+1})$, \ie, the probability distribution of each action $a$ when the agent transitions from state $s_t$ to $s_{t+1}$. 
While we assume the environment is an MDP, in imitation learning the agent has no access to an explicit reward signal.
The actions performed by the teacher are unknown, so we want to find an imitation policy from a set of state-only demonstrations of the teacher $\mathcal{T} = \{\zeta_1, \zeta_2, \ldots, \zeta_N\}$, where $\zeta$ is a state-only trajectory $\{s_0, s_1, \ldots, s_N\}$.

A classic self-supervised IL approach in the LfO area focuses on using two models to learn to imitate.
The models are the Inverse Dynamic Model ($\mathcal{M}$) and the policy model $\pi_\theta$.
The Inverse Dynamic Model is responsible for learning to predict $P(a \mid s_t, s_{t+1})$, which action caused the transition between a given pair of states ($s_t,s_{t+1}$).
After training the ($\mathcal{M}$) model, it is now possible to predict the most possible action taken by the teachers in their collected trajectories that will be used in the policy $\pi_\theta$ training~\cite{TorabiEtAl2018,MonteiroEtAl2020abco}.
Using the pseudo-labels $\hat{a}$ to replace the potential actions that the teachers might have taken builds a natural exploration for LfO methods which helps the agent acquire the capability of generalisation by accessing different states from the original trajectory and acquiring knowledge about state transitions~\cite{gavenski2021self}.
A learner can iterate over this self-supervised pipeline to reduce the error coming from both $\mathcal{M}$ and $\pi_\theta$ models.

LfO strategies offer a more data-efficient approach to imitation learning, since they can learn similar policies to other methods that require labelled data.
Specifically, we consider policies to be similar not only in terms of similar returns, but also policies that generate similar trajectories in the state space. 
This is in contrast to the literature in imitation learning, which often measures similarities by looking at the returns alone for any given task~\cite{TorabiEtAl2018,MonteiroEtAl2020abco,gavenski2020imitating}.
However, just because an agent achieves similar returns in a given task, this does not mean the agent is actually imitating observed behaviour if the policies generate radically different trajectories.

\section{\method}
\label{sec:sail}

\begin{figure*}
    \centering
    \includegraphics[width=0.85\textwidth]{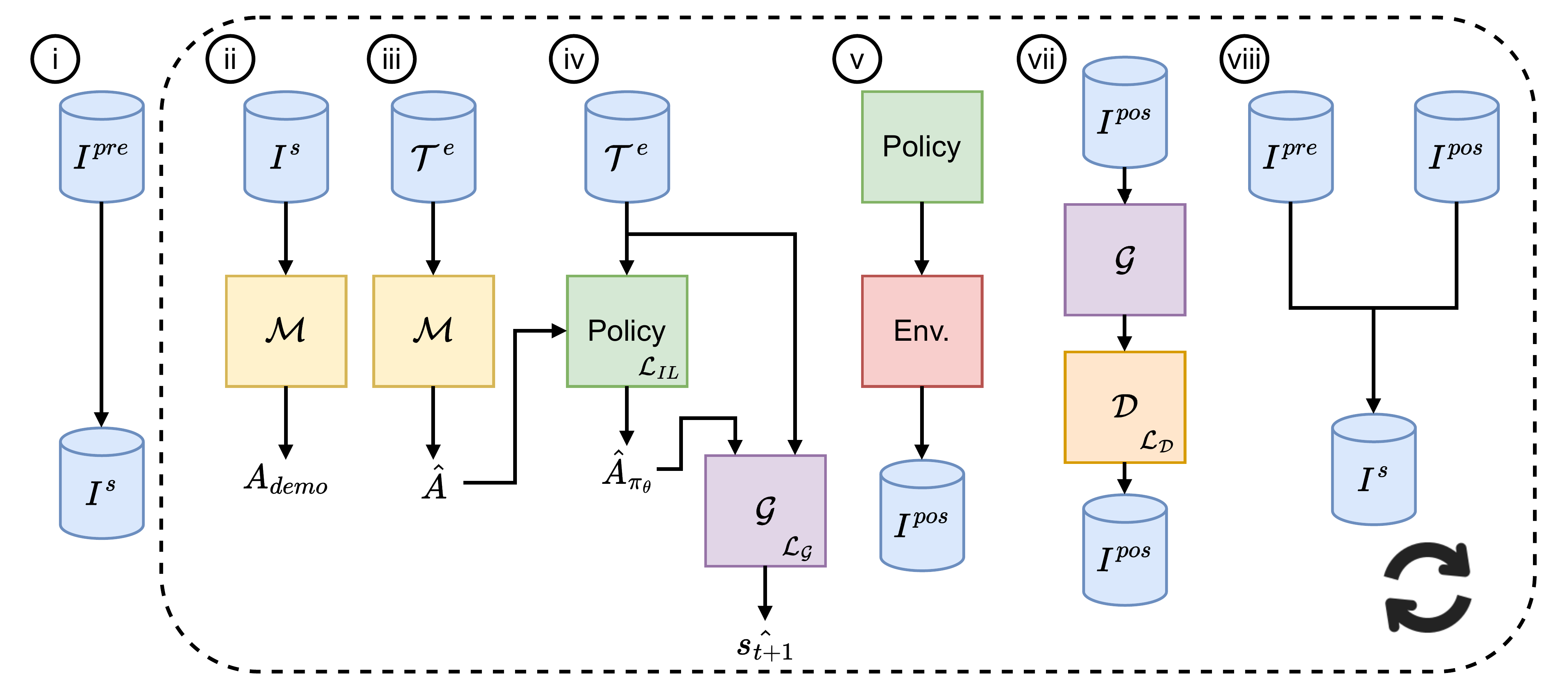}
    \caption{\method~(\abbrev) training pipeline proposed in this paper. All models are initialised with random weights, and the agent interacts with the environments to collect data, so $\mathcal{M}$ learns state transitions. 
    Afterwards, given state-only demonstrations, the $\mathcal{M}$ creates self-supervised labels to use with its policy in a supervised manner. 
    The updated policy then interacts with the environment and collects new samples to be used in a new iteration by the state transition model. 
    Unlike previous work, \abbrev~uses a discriminator to classify whether samples should be appended to the initial samples and a generative model to update its policy to act more like its observation samples. \abbrev~repeats steps (ii) from (viii) for a specific amount of epochs or until both models stop improving.}
    \label{fig:method}
\end{figure*}

 In this paper, we create the \method~(\abbrev), an IL method that interleaves self-supervised and adversarial learning to create a policy based on observation of a teacher without any use of labelled snapshots.
 \abbrev~uses an exploration mechanism based on previous work~\cite{gavenski2020imitating,kidambi2021mobile} to explore when it is unsure of the teacher's actions, and a discriminator model to classify whether the policy trajectory is similar to a teachers' one.
 \abbrev~comprises four different models: 
 \begin{enumerate*}[label=(\roman*)]
    \item $\mathcal{M}$, which predicts an action given a state transition $P(\hat{a} \mid s_t, s_{t+1})$;  
    \item a policy model $\pi_\theta$ that uses the self-supervised labels $\hat{a}$ to mimic a teacher, given a state $P(\hat{a} \mid s_t)$;
    \item a generative model $\mathcal{G}$ conditioned by prediction from $\pi_\theta$ and the current state $P(s_{t+1} \mid s_t, \tilde{a})$; and
    \item a discriminator model $\mathcal{D}$ to discriminate $\pi_e$ and $\pi_\theta$, creating better samples for $\mathcal{M}$ and updating % $\pi_\theta$ 
    weights $\theta$ when the policy is not similar to its proficient counterpart.
 \end{enumerate*}

\begin{algorithm}[!bp]
    \caption{\abbrev}
    \label{algo:method}
    \begin{algorithmic}[1]
        \State Initialise model $\mathcal{M}_\theta$ as a random approximator \label{alg:line:initialize}
        \State Initialise policy $\pi_\theta$ with random weights
        \State Initialise generative model $\mathcal{G}$ with random weights
        \State Initialise discriminator $\mathcal{D}$ with random weights \label{alg:line:finalize}
        \State Generate $I^s$ using $\pi_\theta$ \label{alg:line:ipre}
        \For { $i \gets 1$ to epochs }
            \State Improve $\mathcal{M}_\theta$ by $\Call{supervised}{I^s}$ \label{alg:line:idm}
            \State Use $\mathcal{M}_\theta$ with $\mathcal{T}^{e}$ to predict $\hat{A}$ \label{alg:line:self}
            \State Improve $\pi_\theta$ and $\mathcal{G}$ by $\Call{behaviouralCloning}{\mathcal{T}^{e}, \hat{A}}$ \label{alg:line:policy}
            \State Use $\pi_\theta$ to solve environments $E$ \label{alg:line:enjoy}
            \State Append samples $I^{pos} \gets (s_t, \tilde{a}_t, s_{t+1})$
            \State Append $I^s \gets \forall i \in I^{pos} \;| \; \mathcal{D}(\mathcal{G}(I^{pos}_i)) = 1$ \label{alg:line:append}
        \EndFor
    \end{algorithmic}
\end{algorithm}

 Algorithm~\ref{algo:method} provides an overview of the learning procedure of \abbrev.
 We first initialise all four models with random weights~(Line~\ref{alg:line:initialize}-\ref{alg:line:finalize}).
 With $\pi_\theta$ randomly initialised, we can use this model as a random agent and collect all samples required (Line~\ref{alg:line:ipre}).
Considering that IL methods have no knowledge about optimal or the teacher's action distributions, these random samples help $\mathcal{M}$ to classify each state transition without any biases since the randomly generated ones are equally distributed. %% FRM - uniformly distributed? How can you ensure this with a neural network? Given the nonliniearity of an NN, it is very hard to be sure what kind of distribution you get there, even if you initialise randomly. Do you have an experiment that shows this equal distribution? If not, perhaps just stop the sentence after you say it is unbiased. NSG - in this case it is because all random neural networks follow a uniform distribution. If we plot all data from the dataset, we would see that. 
Once all samples consisting of $(s_t, a, s_{t+1})$ tuples are appended to $I^s$, \abbrev~trains its inverse dynamic model in a supervised manner~(Function~$\Call{supervised}{}$ in Line~\ref{alg:line:idm}).
With its updated weights $\theta$, $\mathcal{M}$ predicts all pseudo-labels $\hat{A}$ to all teacher's transitions in $\mathcal{T}^e$~(Line~\ref{alg:line:self}). 
As the model's weights might not be optimal in each iteration, \abbrev~implements an exploration mechanism that allows for $\mathcal{M}_\theta$ to deviate from its MAP by sampling from its predictions using softmax distributions of the same output as weights.
This mechanism allows \abbrev~to dynamically explore other labels when unsure (with more uniform distributed MAP values) and exploit once its MAPS values are farther apart. 
After creating all self-supervised labels, we train $\pi_\theta$ using a behavioural cloning approach~(Function~$\Call{behaviouralCloning}{}$ in Line~\ref{alg:line:policy}).
Unlike other behavioural cloning approaches, \abbrev~also uses a generative model $\mathcal{G}$ to predict its next state, conditioned by the action predicted by $\pi_\theta$.
Hence, it also updates $\mathcal{G}$ during $\pi_\theta$'s training (we further explain the learning process of the generative model in Section~\ref{sec:sub:generative}).
Afterwards, \abbrev~uses $\pi_\theta$ to create new samples that might help $\mathcal{M}$ better approximate the unknown ground-truth actions from $\mathcal{T}^e$~(Line~\ref{alg:line:enjoy}).
Finally, \abbrev~appends to $I^s$ all samples that $\mathcal{D}$ could not differentiate between $\mathcal{T}^e$ and $\mathcal{G}(I^{pos})$~(Line~\ref{alg:line:append}).
We aim to discard trajectories that could result in $\mathcal{M}$ getting stuck in bad local minima and update $\pi_\theta$ to correct some behaviours that $\mathcal{D}$ uses to differentiate between teacher and student. 
We better explain how $\mathcal{D}$ benefits \abbrev~in Section~\ref{sec:sub:adversarial}.

\subsection{Goal-aware function} \label{sec:sub:adversarial}

Developing a goal-aware function can be a complex task.
Environments in the agent literature have different meanings for what a goal is.
Environments typically have one of two different types of tasks:
\begin{enumerate*}
    \item maintenance; or
    \item achievement~\cite[Chapter 2]{wooldridge2009introduction}.
\end{enumerate*}
Environments with an achievement task define a clear end goal, such as MountainCar~\cite{Moore90efficientmemory-based} -- where an agent has to reach a flag located on top of a mountain with an accumulated reward~$\geqslant -110$.
Since agents in an IL context have no access to the reward signal, we must consider the number of steps an agent performs before reaching its objective.
As environments grow in complexity, such a function will become harder to encode.
By contrast, maintenance task environments usually define a set of states that an agent should not reach.
For example, Ant~\cite{schulman2015high}, where an agent walks as far as possible without reaching angles that it classifies as `falling'.
While others, such as CartPole~\cite{barto1983neuronlike}, define a stopping criterion, \eg, when its pole reaches a certain angle, and an optimal threshold, \eg, maintain its task for 195 consecutive steps.
Thus, encoding a goal-aware function creates a degree of unwanted complexity in a learning algorithm.

By only using samples that reach a goal, \abbrev~acquires examples that have `some' degree of optimality, and approximates $I$'s samples from $\mathcal{T}^e$~\cite{gavenski2020imitating}.
Nevertheless, classifying whether samples are close to $\mathcal{T}^e$ might be difficult.
First, defining what indicates a sample being close to a proficient teacher is hard.
If we consider a stationary agent (which \abbrev~is), we might discard samples that allow $\mathcal{M}$ to accurately predict transitions due to their distance to $\mathcal{T}^e$ states alone.
Therefore, to achieve better policies, \abbrev~needs a goal-aware function that allows for $\mathcal{M}$ to deal with sub-optimal samples.

To remove the usage of hand-crafted goal-aware functions, \abbrev~uses a discriminator $\mathcal{D}$ to discriminate between $\pi_\theta$ and $\pi_e$.
By doing so, \abbrev~eliminates all human intervention and gains a non-greedy sampling mechanism by using a model to classify which agent created a trajectory. 
Moreover, since $\mathcal{D}$ starts with random weights~(Line~\ref{alg:line:finalize}), it allows samples that did not reach a `goal', \ie, sub-optimal samples, to be appended to $I^s$.
However, considering that \abbrev~works under LfO constraints (not having access to $\pi_e$ actions), there is a need to create a mechanism that can discriminate between teachers' and students' state-only trajectories.
Using state-only trajectories from $\pi_\theta$ and $\pi_e$, and an adversarial learning approach~(Equation~\ref{eq:adversarial}), \abbrev~allows $\pi_\theta$ via $\mathcal{G}$ to be updated from its gradient flow from $\mathcal{D}$.

\begin{equation} \label{eq:adversarial}
    \begin{split}
        \min_{\mathcal{M} \cup \pi} \max_{\mathcal{D}} &\abbrev(\mathcal{M}, \pi, \mathcal{G}, \mathcal{D}) = \\
        &\mathbb{E}_{r \sim R(\pi_\epsilon,Env)}[log(\mathcal{D}(r))] + \\
        &\mathbb{E}_{r' \sim R(\pi_\theta, Env)}[log(1 - \mathcal{D}(\mathcal{G}(r', \pi_\theta(r'))))]      
    \end{split}
\end{equation}

\noindent
Considering that \abbrev~uses a few samples (in the form of $\mathcal{T}^e \cup \mathcal{T}^\pi$), it is important to avoid $\mathcal{D}$ overfitting. 
For that, we record all $\pi_\theta$ trajectories in a replay buffer $RB$ and only sample a few trajectories from both sets at each iteration, $RB_{(s_t, \cdots, s_{t+n})} \sim \mathcal{T}^e \cup \mathcal{T}^{\pi}$.
By only using a sub-set from each sample pool of trajectories, \abbrev~avoids overfitting $\mathcal{D}$ during first iterations (where $\pi_\theta$'s trajectories are considerably different).

\subsection{Generative model} \label{sec:sub:generative}

\abbrev~uses a generative model in two different steps.
Firstly during Function~$\Call{behavioralCloning}{}$; and afterwards, when selecting which samples should be appended to $I^s$
Although \abbrev~adds a new model to its pipeline, in these situations, it benefits in two folds:
\begin{enumerate*}[label=(\roman*)]
    \item intrinsically encodes environments physics in $\pi_\theta$; and
    \item it updates $\pi_\theta$ when using $\mathcal{D}$ via its gradient flow.
\end{enumerate*}
Which we believe far surpasses the overhead cost created by using a generative model.

The first benefit results from \abbrev~updating $\mathcal{G}$ weights using Equation~\ref{eq:generative}.
Thus, when using the generative model in Line~\ref{alg:line:policy}, \abbrev~allows $\mathcal{L}_{\mathcal{G}}$ to update $\pi_\theta$ to create actions that would correctly condition $\mathcal{G}$ to generate correct state transitions and equal to those observed.
Moreover, by learning how to properly decode from $(s_t, \tilde{a_t}) \mapsto \hat{s_{t+1}}$, the generative model becomes a forward dynamics model, which helps $\pi_\theta$ encode some of the environments' dynamics, \eg, physics.
We hypothesise that it is also possible only to update $\mathcal{G}$ weights when $\pi_\theta$ correctly predicts a self-supervised label.
However, this creates two different problems.
The first problem is that not always $\mathcal{M}$ will be correct.
Thus, $\pi_\theta$ being correct about $\hat{a}$ might not be indicative of how accurate it is in conditioning $\mathcal{G}$.
The second problem originates from the fact that $\mathcal{M}$ might stop predicting some actions when being stuck in local minima~\cite{gavenski2020imitating}.
In this scenario, $\mathcal{G}$ will not update for state transitions for the action not being predicted.
Therefore, $\mathcal{G}$ will update $\pi_\theta$ fewer times, resulting in less exploration, since updating $\pi_\theta$ weights with two different objectives can also help it to no be stuck at local minima.

\begin{equation} \label{eq:generative}
    \mathcal{L}_{\mathcal{G}} = -\frac{1}{N} \left [\sum_{i=1}^{N} s_{i+1} \cdot log(\mathcal{G}(s_i, \pi_\theta(s_i)) \right ]
\end{equation}

The second benefit originates from the fact that \abbrev~has an adversarial training mechanism in its pipeline.
Hence, $\mathcal{D}$ directly updates $\pi_\theta$ weights via gradient flow when it correctly discriminates between teacher and student (Equation~\ref{eq:adversarial}).
This behaviour is beneficial because updating $\pi_\theta$ through $\mathcal{D}$ allows for the agent to have a direct temporal signal, \ie, where it deviates from its teachers' observations.
Consequently, since \abbrev~maintains all original behavioural cloning techniques, it creates agents that mimic teachers' trajectories more accurately while not losing performance.

Both generative update moments allow \abbrev~to have more precise trajectories (further explored in Section~\ref{sec:results}), as well as more trajectories that are similar to their source (discussed in Section~\ref{sec:sub:mimic}).
We believe having trajectories closer to the teachers is beneficial since it avoids unwanted biases due to previous methods only using the performance metrics, which does not carry behaviour meaning~\cite{Gavenski2022how}.

\section{Experimental Results} \label{sec:results}

We tested \abbrev~and all baselines described in Section~\ref{sec:related-work} with four different environments:
\begin{enumerate*}[label=(\roman*)]
    \item CartPole;
    \item MountainCar;
    \item Acrobot; and
    \item LunarLander.
\end{enumerate*}
We use OpenAI Gym~\cite{brockman2016openai} versions for all environments.
Figure~\ref{fig:envs} illustrates a single frame for each of these environments, while Section~\ref{sec:sub:implementation} gives a brief description of all environments and \abbrev~neural network topology.

\subsection{Implementation and Metrics} \label{sec:sub:implementation}

We follow Gavenski~\etal~\cite{gavenski2020imitating} implementation for our agents.
Therefore, $\pi_\theta$ is a Multi-Layer Perceptron (MLP) model with $2$ hidden layers with $32$ neurons and $2$ self-attention modules after each layer.
$\mathcal{M}$ is an MLP with $2$ hidden layers with $32$ neurons, $2$ self-attention modules and $2$ Layer Normalisation layers.
$\mathcal{G}$ is an MLP with $2$ hidden layers with $2 \times (\mid s \mid + 1)$ neurons, where $\mid s \mid$ is the size of the environment state vector, and no self-attention or normalisation layers.
$\mathcal{D}$ is a Long Short Term Memory~\cite{graves2012long} with $2$ layers, $32$ neurons each, and dropout of $50\%$.
The official \abbrev~implementation can be found at: \url{https://github.com/NathanGavenski/SAIL}.

To measure our experiments, we will use two main metrics: 
Average Episodic Reward (AER), and \textit{Performance} ($\mathcal{P}$)~\cite{HoErmon2016} metrics.
AER is the average of all accumulated rewards for a consecutive amount of tries in each environment.
Performance is returned by calculating the average reward for each run scaled to be within $[0,1]$, where zero is a behaviour compatible with a random policy ($\pi_\xi$) reward, and one a behaviour compatible with the teacher ($\pi_\varepsilon$).
\begin{equation} \label{eq:performance}
    \mathbb{P} = \frac{\sum_{i=1}^{E}\frac{\pi_\phi(e_i)-\pi_\xi(e_i)}{\pi_\varepsilon(e_i) - \pi_\xi(e_i)}}{E}
\end{equation}
It is possible for a model to achieve scores $< 0$ if it has the worst performance than a random policy and $> 1$ if the model can perform better than its teacher.
We do not use accuracy for evaluation since achieving high accuracy in Imitation Learning tasks does not guarantee good results in solving a task.

We now briefly describe all environments used in this work.
\begin{itemize}

\item[$\bullet$] \textbf{CartPole-v1} is an environment where an agent moves a car sideways, applying force to a single pole. 
The goal is to prevent the pole from falling over. 
The space state has four dimensions:\textit{car position}, \textit{car velocity}, \textit{pole angle}, and \textit{pole velocity} at tips. 
The agent receives a single reward point every time the pole remains upright. 
Barto~\cite{barto1983neuronlike} describes solving CartPole as getting an average reward of $195$ over $100$ consecutive trials.

\item[$\bullet$] \textbf{MountainCar-v0} environment consists of a car situated in a valley. 
The agent needs to learn to leverage potential energy by driving up the opposite hill until completing the goal. 
The state-space has two continuous attributes: \textit{velocity} and \textit{position} and three discrete action spaces: \textit{left}, \textit{neutral}, and \textit{right}. 
A reward of $-1$ is provided for every time step until the goal position of $0.5$ is reached. 
The first state starts in a random position with no velocity. 
Moore~\cite{Moore90efficientmemory-based} defines solving MountainCar as getting an average reward of $-110$ over $100$ consecutive trials.

\item[$\bullet$] \textbf{Acrobot-v1}, based on Sutton's work~\cite{SuttonAcrobot}, is an environment where an agent has two joints and two links. 
The joint between the two links is actuated. 
The state space consists of: {$\{\cos\theta_1, \sin\theta_1,$ $\cos\theta_2, \sin\theta_2, \theta_1, \theta_2\}$}, and the action space consists of the $3$ possible forces. 
The goal is to move the end of the lower link up to a given height. 
Although Acrobot is an achievement task environment, it does not have a specified reward threshold. 

\item[$\bullet$] \textbf{LunarLander-v2}, created by Klimov~\cite{brockman2016openai}, is an environment where an agent needs to land on the moon under low gravity conditions. 
The state space is continuous, and the action space is discrete. 
There are four actions: \textit{do nothing}, \textit{move left}, \textit{right}, and \textit{reduce the falling velocity}. 
All actions have a reward of $-1$, except for \textit{do nothing} state, which is $-0.3$.
A positive value is returned when the agent moves in the right direction (always at $0,0$ coordinates).
LunarLander-v2 is solved when the agent receives a reward of $200$ over $100$ constitutive trials.

\end{itemize}

\begin{table*}[!htp]
    \centering
    \caption{\abbrev~and baselines results for all environments.}
    \label{tab:results}
    \begin{tabular*}{\textwidth}{l@{\extracolsep{\fill}}crrrr}
        \toprule
            \multicolumn{1}{c}{Algorithm} & Metric & CartPole & MountainCar & Acrobot & LunarLander  \\
         
         \midrule
         \midrule 
            
            \multirow{2}{*}{Random} & AER           &  $21,92 \pm $ & $-200 \pm 0$      & $-499.36 \pm $    & $-170.47 \pm $        \\
                                    & $\mathcal{P}$ & $0$           & $0$               & $0$               & $0$                   \\ \midrule
            \multirow{2}{*}{Expert} & AER           & $500 \pm 0 $  & $-98.03 \pm 8.17$ & $-74.85 \pm 8.61$ & $256.79 \pm 21.38$    \\
                                    & $\mathcal{P}$ & $1$           & $1$               & $1$               & $1$                   \\

        \midrule
        \midrule

            \multirow{2}{*}{BC}      & AER           & $218.53 \pm 160.71$  & $-102.06 \pm 4.23$           & $-80.21 \pm 3.61$             & $63.05 \pm 79.50$              \\
                                     & $\mathcal{P}$ & $0.37$               & $0.97$                       & $0.99$                        & $0.63$                         \\ \midrule
            \multirow{2}{*}{GAIL}    & AER           & $302.03 \pm 158.96$  & $-200 \pm 0$                 & $-274.27 \pm 116.85$          & $120.21 \pm 28.03$             \\
                                     & $\mathcal{P}$ & $0.41$               & $0$                          & $0.54$                        & $0.66$                         \\ \midrule
            \multirow{2}{*}{GAIfO}   & AER           & $\mathbf{500 \pm 0}$ & $-200 \pm 0$                 & $-128.20 \pm 15.88$           & $\mathbf{200 \pm 29.95}$       \\
                                     & $\mathcal{P}$ & $\mathbf{1}$         & $0$                          & $0.85$                        & $\mathbf{0.86}$                \\ \midrule
            \multirow{2}{*}{IUPE}    & AER           & $\mathbf{500 \pm 0}$ & $-166.97 \pm 18.34$          & $\mathbf{-75.65 \pm 12.85}$   & $-81.34 \pm 74.5$              \\
                                     & $\mathcal{P}$ & $\mathbf{1}$         & $0.32$                       & $\mathbf{1}$                  & $0.21$                         \\ \midrule
            \multirow{2}{*}{\abbrev} & AER           & $\mathbf{500 \pm 0}$ & $\mathbf{-99.35 \pm 1.84}$   & $-78.84 \pm 0.41$             & $183.62 \pm 5.63$              \\
                                     & $\mathcal{P}$ & $\mathbf{1}$         & $\mathbf{0.99}$              & $0.99$                        & $0.83$                         \\

        \bottomrule
    \end{tabular*}
\end{table*}

\subsection{Results}

Table~\ref{tab:results} shows  results for all baselines and \abbrev~in four different environments.
\abbrev's performance is closer to the teacher's reward in almost all environments (CartPole, MountainCar and Acrobot), and it performs worst in the LunarLander environment, resulting in the best algorithm throughout all environments.
When comparing \abbrev~to other methods, we observe it yields a lower standard deviation ($\leqslant 1$) in the first three environments, with a higher deviation for the LunarLander environment ($5.63$).
We believe these lower deviations are due to the gradient flow from $\mathcal{D}$ into $\pi_\theta$ discussed in Section~\ref{sec:sub:adversarial}, and $\mathcal{G}$ intrinsically encoding each environment physics into $\pi_\theta$ without a direct signal (Section~\ref{sec:sub:generative}).
By properly encoding physics in its policy, \abbrev~achieves a behaviour that helps its agent yield similar results to the teacher since it has knowledge on how $s_{t+1}$ should be given $a$ and $s_t$.
Moreover, $\mathcal{D}$ allows $\pi_\theta$ to have a temporal signal in its trajectory, helping to correct any unwanted/divergent behaviour that helps $\mathcal{D}$ discriminate against teacher and student.

Conversely, \abbrev~does not achieve the best results for Acrobot and LunarLander.
For the Acrobot environment, \abbrev~achieves an accumulated reward lower than IUPE ($3.19$ lower).
However, its standard deviation is even lower than behavioural cloning, which had labelled snapshots during its training.
We believe this is the case for this environment because Acrobot rewards hectic behaviour from the agent, which IUPE deeply beneficiates given its exploration mechanism, while $\mathcal{D}$ incentives \abbrev~to have more consistent trajectories.
We hypothesise that for cases where \abbrev~has a higher exploration ratio, reducing the gradient from its adversarial phase would be beneficial, avoiding it getting into an exploitative phase too soon.
As for the LunarLander environment, \abbrev~achieves a similar result to GAIfO, which had $16.38$ more accumulated reward, but with $24.32$ more deviation points.
We believe this is the case for \abbrev~due to it learning the proficient behaviour much more closely than the other IL counterparts.
If we compare BC's performance, we observe that by having the finer-grained information of all actions, the results dramatically changed -- which we draw the comparison to \abbrev's results.
For the LunarLander environment, IUPE has the worst and only negative result.
We believe that such a result originates from the fact that LunarLander optimal behaviour highly correlates to the agent and goal initialisation, which IUPE lacks mechanisms to understand from its proficient~source.

Finally, we observe that for environments with more relation between states (\ie, carrying momentum), such as MountainCar, \abbrev~performs best.
While most baselines yielded policies closer or equal to a random one, \abbrev~achieved a performance $\approx 1$.
During experimentation, we observe that most other methods require the agent to be in a specific state, \ie, stopped or with almost no movement force.
However, we did not notice a similar behaviour since \abbrev~receives information from its discriminator model.

\section{Discussion} \label{sec:discussion}

In this section, we consider the following:
\begin{enumerate*}[label=(\roman*)]
    \item how \abbrev~learns with different amounts of samples; and
    \item how $\pi_\theta$ behaves compared to its performance and discriminator accuracy.
\end{enumerate*}
The first case allows us to understand the trade-off between having more or fewer trajectories than those presented in Section~\ref{sec:results}.
Understanding this balance is essential, so \abbrev~learns with as fewer samples as possible and, therefore, faster.
We investigate the second case to understand how much $\pi_\theta$ approximates from $\pi_e$ trajectories.
IL works use performance and AER as metrics but do not consider how similar the policy is to its teacher.

\subsection{Sample Efficiency} \label{sec:sub:efficiency}

% Env	         Expert	   std     	Random
% CartPole	     500,00	   0,00  	21,92
% MountainCar	 -98,03	   8,17	   -200,00
% Acrobot	     -74,85	   8,61	   -499,36
% LunarLander	 256,79    21,38   -170,47

\begin{table*}[!tp]
    \centering
    \caption{Results for \abbrev~with different sample sizes.}
    \label{tab:efficiency}
    \begin{tabular*}{\textwidth}{l@{\extracolsep{\fill}}rrrrrr}
        \toprule
        \multicolumn{1}{l}{Environment}   & 
        \multicolumn{1}{c}{Trajectories} & 
        \multicolumn{1}{c}{$\mathcal{P}$} & 
        \multicolumn{1}{c}{$AER~(avg)$}   & 
        \multicolumn{1}{c}{$AER~(\min)$}  &
        \multicolumn{1}{c}{$AER~(\max)$}  &
        \multicolumn{1}{c}{$SD$}         \\
        
        \midrule 
        
        % CARTPOLE
        \multirow{5}{*}{CartPole} 
            % 1 SAMPLE
            & $1$   & $0.55$ & $1.55$ & $86.23$ & $457.30$ & $\pm133.98$ \\
            % 25 SAMPLES
            & $25$  & $1$ & $500$ & $500$ & $500$ & $\pm0$ \\
            % 50 SAMPLES
            & $50$  & $1$ & $500$ & $500$ & $500$ & $\pm0$ \\
            % 75 SAMPLES
            & $75$  & $1$ & $500$ & $500$ & $500$ & $\pm0$ \\
            % 100 SAMPLES
            & $100$ & $1$ & $500$ & $500$ & $500$ & $\pm0$ \\
        \midrule

        % MOUNTAIN-CAR
        \multirow{5}{*}{MountainCar} 
            % 1 SAMPLE
            & $1$   & $0$ & $-200$ & $-200$ & $-200$ & $\pm0$ \\
            % 25 SAMPLES
            & $25$  & $0.87$ & $-109.96$ & $-114.40$ & $-103.60$ & $\pm4.21$ \\
            % 50 SAMPLES
            & $50$  & $0.96$ & $-101.78$ & $-103.70$ & $-99.11$ & $\pm2.02$ \\
            % 75 SAMPLES
            & $75$  & $0.99$ & $-99.35$ & $-102.10$ & $-97.38$ & $\pm1.84$ \\
            % 100 SAMPLES
            & $100$ & $0.98$ & $-101.31$ & $-109.90$ & $-97.87$ & $\pm4.93$ \\
        \midrule

        % ACROBOT
        \multirow{5}{*}{Acrobot} 
            % 1 SAMPLE
            & $1$   & $0.98$ & $-87.47$ & $-112.20$ & $-77.75$ & $\pm14.03$ \\
            % 25 SAMPLES
            & $25$  & $0.99$ & $-77.95$ & $-79.75$ & $-76.78$ & $\pm1.16$ \\
            % 50 SAMPLES
            & $50$  & $0.99$ & $-79.33$ & $-80.33$ & $-78.23$ & $\pm0.96$ \\
            % 75 SAMPLES
            & $75$  & $0.99$ & $-78.84$ & $-79.46$ & $-78.46$ & $\pm0.41$ \\
            % 100 SAMPLES
            & $100$ & $0.99$ & $-79.36$ & $-80.72$ & $-76.43$ & $\pm1.73$ \\
        \midrule

        % LunarLander
        \multirow{5}{*}{LunarLander} 
            % 1 SAMPLE
            & $1$   & $0.31$ & $-30.13$ & $-76.95$ & $55.28$ & $\pm52.28$ \\
            % 25 SAMPLES
            & $25$  & $0.86$ & $196.50$ & $148$ & $242.20$ & $\pm36.86$ \\
            % 50 SAMPLES
            & $50$  & $0.83$ & $133.19$ & $-24.84$ & $207.30$ & $\pm98.04$ \\
            % 75 SAMPLES
            & $75$  & $0.83$ & $183.62$ & $175.20$ & $188.90$ & $\pm5.63$ \\
            % 100 SAMPLES
            & $100$ & $0.81$ & $151.26$ & $10.72$ & $204.50$ & $\pm79.45$ \\        
        \bottomrule
    \end{tabular*}
\end{table*}
We observe that \abbrev~inherits two different behaviours from behavioural cloning methods.
Firstly, it requires a sample size bigger than a single trajectory to learn how to achieve a performance similar to its teacher.
This is no surprise since $1$ trajectory barely provides any information for \abbrev~to learn how to encode different states and generalise significantly in each environment~\cite{hoangLe2018}.
The second behaviour \abbrev~inherits is it fails to scale according to the number of samples due to compounding error~\cite{swamy2021moments}.
As the samples grow, the policy diverges less from its observed trajectories, decreasing the agent's performance.
Therefore, behavioural cloning methods need to find the correct number of trajectories they should use. 
Conversely, as Table~\ref{tab:efficiency} shows, \abbrev~achieves a performance close to $1$ with $25$ episodes, only decreasing its standard deviation for each row consecutively and increasing again with $100$ trajectories.
We believe that \abbrev~achieves this result due to the updates from $\mathcal{G}$ and, therefore, $\mathcal{D}$.
Using a different objective, \eg, recreating trajectories closer to the teachers', \abbrev~results in fewer episodes needed than previous methods would because $\pi_\theta$ is less dependent on its LfO objective.

Nevertheless, \abbrev~achieves these results also due to using a smaller discriminator model, which comes with the price of being sequential.
We hypothesise that to keep the sample size small, \abbrev~cannot use a larger sequence model, such as Transformers~\cite{vaswani2017attention}, since the high number of parameters means that the discriminator model would easily overfit on its small dataset.
We believe that increasing \abbrev~efficiency would require further experimentation regarding the augmentation of observations since performing modification of teacher samples requires prior domain knowledge so as not to augment tuples into impossible or undesired transitions.

\begin{table}[!bp]
    \centering
    \caption{Discriminator's accuracy ($\mathcal{D}$), Generator's loss ($\mathcal{G}$) and Policy's performance ($\pi_\theta$) for each environment.}
    \label{tab:mimic}
    \begin{tabular*}{\columnwidth}{l@{\extracolsep{\fill}}rrr}
        \toprule
            Environment & $\mathcal{D}$'s Accuracy (\%)
            & $\mathcal{G}$'s Loss & $\pi_\theta$'s Performance \\
        \midrule
             \multirow{1}{*}{CartPole} & 
            $49.44 \pm 1.12$ & 0.0015 & 1 \\ 
             \midrule
             
             \multirow{1}{*}{MountainCar} & 
             $49.76 \pm 0.83$  & 0.0029 & 0.98 \\ 
             \midrule
             
             \multirow{1}{*}{Acrobot} & 
             $50.13 \pm 3.21$ & 0.8685 & 0.99 \\ 
             \midrule
             
             \multirow{1}{*}{LunarLander} & 
             $49.08 \pm 1.28$ & 0.0308 & 0.81 \\
             
        \bottomrule
    \end{tabular*}
\end{table}

\subsection{Imitation Behaviour} \label{sec:sub:mimic}

IL metrics usually rely only on the accumulated reward of the agent to judge how well a policy learned to mimic its teacher's behaviour.
Most works, such as this, use $\mathcal{P}$ as a metric, which, we believe, fails to show IL agents' intricacies (such as trajectory).
By only using the reward signal to evaluate these agents, we can not conclude that the agent behaves like its teacher.
There might be a proficient behaviour desired in an environment that is not encoded in the reward function or a stochastic behaviour not apparent in its observation. 
Therefore, $\mathcal{P}$ may fail to measure a divergence in trajectory since the accumulated reward might be equal.
Let us use a maze environment as an example, where there are always two trajectories with equal lengths, but in one path, there is a slight chance of the floor breaking.
In this case, a conservative agent might never take the trajectory with the faulty floor.
On the other hand, a more aggressive agent will not consider this when making its way through the maze.
An imitation learning agent that learns from the conservative policy might inherit the bias of never stepping into the floor that might break. 
However, the performance will not measure this behaviour, even if the student becomes closer to the aggressive policy. 

To avoid this issue, we analyse not only $\pi_\theta$'s performance but also $\mathcal{D}$ accuracy and $\mathcal{G}$ error.
If $\pi_\theta$ achieves higher performance, while $\mathcal{D}$ has high accuracy and $\mathcal{G}$ has a lower error, it means that although $\pi_\theta$ correctly encoded proficient behaviour, it has a trajectory that is not close from its teacher.
Moreover, if $\pi_\theta$ achieves a performance close to $1$, and $\mathcal{D}$ has lower accuracy and $\mathcal{G}$ has a higher error, it means that $\pi_\theta$ yields the same reward as its teacher, but its generator learned how to encode next states to `fool' $\mathcal{D}$, but does not follow the observations correctly. 
Therefore, we are looking for a scenario where $\pi_\theta$ has $\mathcal{P} \approx 1$, $\mathcal{D}$ an accuracy $\approx 50\%$ and a lower error for its $\mathcal{G}$ during $\Call{behaviouralClonning}{}$ (Line~\ref{alg:line:policy} in Algorithm~\ref{algo:method}.
Table~\ref{tab:mimic} shows all three metrics results for all four environments used in this work.

\begin{figure}[!bp]
    \centering
    \includegraphics[width=\columnwidth]{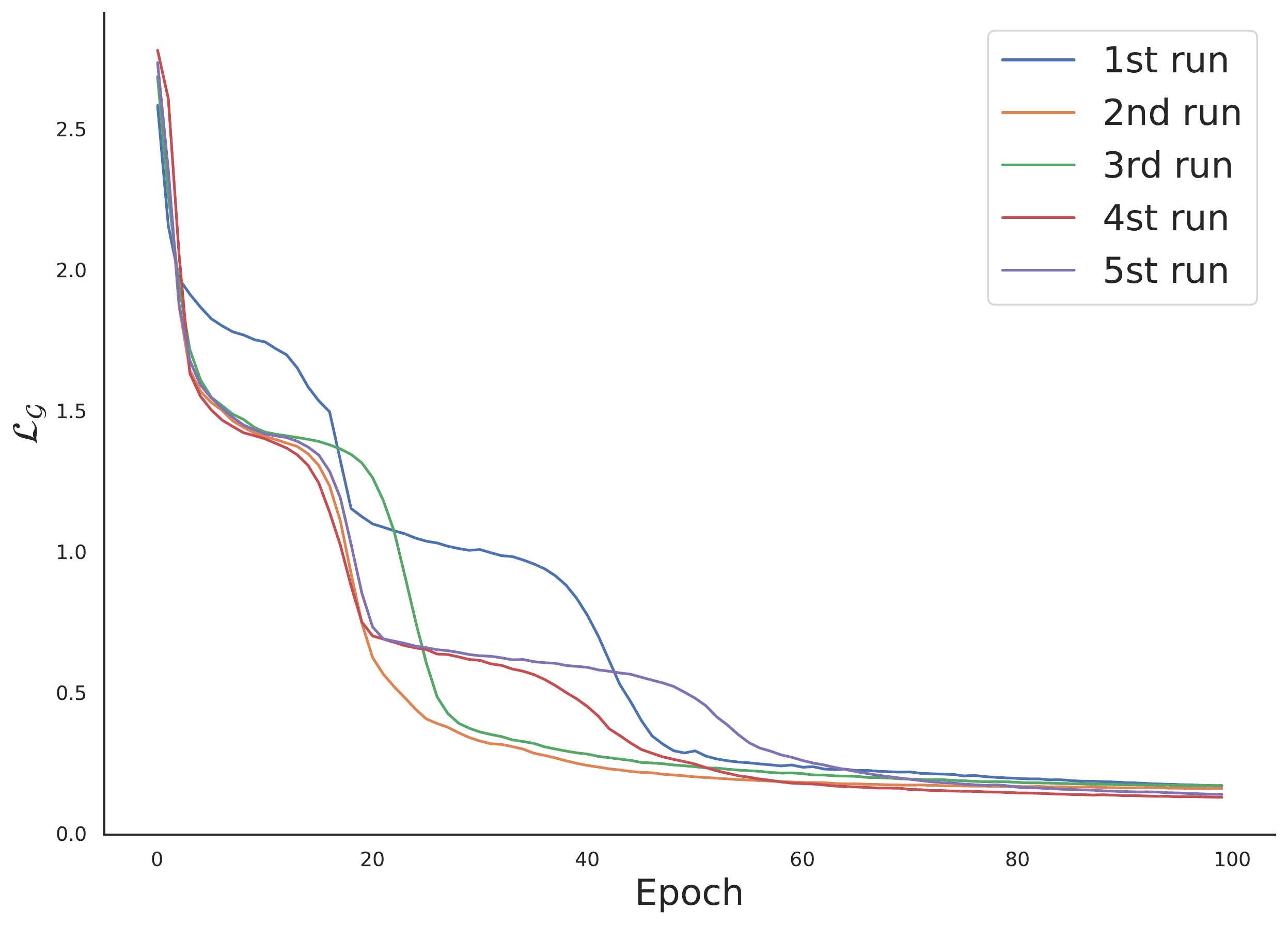}
    \caption{Error rate for $5$ different runs in Acrobot environment.}
    \label{fig:acrobot}
\end{figure}

We observe that for all environments, when $\pi_\theta$ yields its best result, $\mathcal{D}$ has results equal to a random model, \eg, it is not able to discriminate whether a trajectory comes from teacher or student.
Additionally, when we compare $\mathcal{G}$ error during its first learning phase, where it tries to recreate the next state from a teacher's trajectory, it has a lower error, meaning it correctly learned how to encode state transitions.
Therefore, \abbrev~yields a policy consistent with proficient rewards while keeping a consistent trajectory with its learned source.
We note that error is highly contextual to each environment state encodings.
For example, Acrobot's states is a vector with $6$ values, which consists of $4$ values varying from $[-1, 1]$ and two values from $[-12.57, 12.57]$ and $[28.27, 28.27]$.
Figure~\ref{fig:acrobot} displays different error margins for $5$ executions of \abbrev.
It is possible to note that, although $\mathcal{G}$ has a higher error rate for the Acrobot environment, its initial value ($\approx 2.5$) is higher due to the environment encoding characteristics.
Thus, its higher error rate (when compared with all other environments), contextually, could be considered a lower margin rate.

\section{Conclusion} \label{sec:conclusion}
In this paper, we developed a novel LfO approach that uses an adversarial module to learn how to imitate the behaviour of teachers without accessing its actions and is capable of achieving state-of-the-art results in performance and efficiency.
We evaluate our model under different amounts of sampled behaviour and compare it with various baselines from the literature. 
\abbrev~(\method) achieves significant results in both Performance and AER with fewer samples for two main reasons. 
First, we use an adversarial mechanism to better approximate our model's behaviour with the teacher's. 
Such a mechanism is connected to our end-to-end model and iteratively learns through a loss error, which makes the model achieve higher returns.
Second, \abbrev~uses an exploration technique that helps the model collect the best data for each interaction, resulting in a model's convergence in fewer steps.
Finally, we evaluate and discuss the metrics by which we can measure how much a policy actually imitates sampled behaviour. 
To that end, we carry out an ablation study in Section~\ref{sec:discussion} where we show that \abbrev~is capable of achieving proficient rewards while still `fooling' its discriminator, including analysis through our obtained results.

In future work, we aim to test our method in a variety of different environments to understand better \abbrev's ability to imitate other agent behaviour vis-à-vis other baselines from literature.
We believe that by changing our model's topology, we can test with environments that represent their states with images, which could result in learning from teachers by collecting available videos from the internet (\eg, YouTube) since we do not need previous knowledge of actions used by the teachers.
And further investigate how we can measure proficient behaviour in different environments for a better understanding of possible emergent behaviour by the agent, and its impact on real-world applications.

\section{Acknowledgements}
This work was supported by UK Research and Innovation [grant number EP/S023356/1], in the UKRI Centre for Doctoral Training in Safe and Trusted Artificial Intelligence (\url{www.safeandtrustedai.org}) and made possible via King's Computational Research, Engineering and Technology Environment (CREATE)~\cite{create}.

\bibliographystyle{IEEEtran}
\bibliography{references}

\end{document}